\documentclass[]{ceurart}

\usepackage{microtype}
\usepackage{booktabs}
\usepackage{multirow}
\usepackage{float}
\usepackage{amsmath}
\usepackage{tablefootnote}
\usepackage{enumitem}
\usepackage{tabularx}
\usepackage[most]{tcolorbox}

\begin{document}

\copyrightyear{2026}
\copyrightclause{Copyright for this paper by its authors.
  Use permitted under Creative Commons License Attribution 4.0
  International (CC BY 4.0).}

\conference{AI for Education Day, SIGKDD 2026, August 12, 2026, Jeju, Korea}

\title{When Rubrics Change: Cross-Rubric Generalization for Critical Thinking Essay Scoring}


\author[1]{Nischal {Ashok Kumar}}[email=nashokkumar@umass.edu]
\author[1]{Payu Wittawatolarn}
\author[1]{Sana Kang}
\author[2]{Marisa C. Peczuh}
\author[3]{Blair Lehman}
\author[4]{Ryan Baker}
\author[2]{Caitlin Mills}
\author[5]{Sherry Lachman}
\author[6]{Ruochen Sun}
\author[1]{Andrew Lan}

\address[1]{University of Massachusetts Amherst, United States}
\address[2]{University of Minnesota, United States}
\address[3]{Brighter Research, United States}
\address[4]{Adelaide University, Australia}
\address[5]{Advanced Education Research and Development Fund (AERDF), United States}
\address[6]{Independent Researcher, United States}

\begin{abstract}
Automated essay scoring (AES) research has largely focused on cross-prompt generalization, where essays from unseen prompts are scored while the scoring criteria are typically held constant. In practice, however, educators may revise or even introduce new rubrics in their scoring task, to evaluate different aspects of essays. We study \emph{cross-rubric generalization}: training on essays labeled under one set of rubrics and evaluating on previously unseen rubrics, which target different aspects of the essay. We use  a Large Language Model (LLM) fine-tuning framework with two components: rubric-agnostic intermediate representations, called \emph{traits}, and target-essay supervision under seen rubrics during training. On an AES dataset augmented with multiple rubric-defined labels of student critical thinking skills, we find that traits improve macro F1 by 5.0\% over a baseline without traits in the hardest setting, where both target rubrics and target essays are unseen during training. We further find that increasing target-essay supervision improves performance, with our best fine-tuned open-source Llama-based model outperforming GPT-5-mini prompting by 2.1\% macro F1 and trailing GPT-5 by 1.9\%. These results show that trait-based intermediate structure and controlled supervision improve generalization to unseen rubrics.

\makebox[0pt][l]{\includegraphics[height=0.4cm]{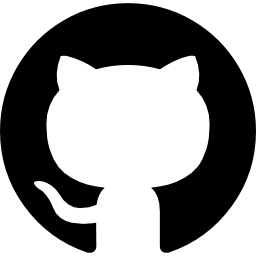}}
\hspace{1em} {\small \texttt{\href{https://github.com/umass-ml4ed/generalization-in-essay-scoring}{github.com/umass-ml4ed/generalization-in-essay-scoring}}}
\end{abstract}


\maketitle

\section{Introduction}

Automated essay scoring (AES) is a long-standing and widely studied problem, motivated by the need to support scalable and consistent assessment of student writing \citep{burstein1999automated, alikaniotis2016automatic, dong2017attention}. Beyond in-domain scoring, a growing line of work has examined how AES systems generalize to essays written in response to previously unseen prompts, where the essay content and topical distributions change but the scoring criteria (\emph{i.e.}, the target rubric) do not \citep{Ridley_He_Dai_Huang_Chen_2021,do-etal-2025-towards,eltanbouly-etal-2025-trates}. This cross-prompt setting has become an important testbed for robust AES, since practical deployments often require models to score writing drawn from new prompts without retraining from scratch for each prompt.

However, real-world assessment settings can require more than generalizing to new prompts. Educators may revise scoring criteria or even introduce new rubrics that evaluate different aspects of the essay; obtaining expert annotation on these new rubrics can be expensive. This challenge is especially relevant in critical thinking assessment, where writing may be evaluated along multiple criteria, each defined by its own rubric, and new rubric criteria may be added over time \citep{peczuh2025toward}. Thus, an interesting problem to study in such settings is \emph{cross-rubric generalization}: training a model on essays labeled under one rubric and applying it to score essays under a previously unseen rubric. Unlike cross-prompt generalization \citep{Ridley_He_Dai_Huang_Chen_2021,do-etal-2025-towards,eltanbouly-etal-2025-trates}, cross-rubric generalization changes the scoring criteria used to evaluate the essay. The model must therefore generalize to a new rubric, where the relevant textual evidence, its interpretation, and the definition of proficiency may differ from those seen during training.

To address this issue, we study cross-rubric generalization within an LLM fine-tuning framework for rubric-based essay scoring from essay text and rubric descriptions. We instantiate this setting on the recent critical thinking essay scoring dataset of \citep{peczuh2025toward}, which contains 500 student essays annotated under six rubric-based evaluation dimensions spanning three broader critical thinking skills. A central component of our approach is a set of rubric-agnostic analytic representations, which we call \emph{traits}, grounded in argumentation and critical thinking theory. These traits capture lower-level properties that recur across the rubrics in this dataset, such as claim explicitness and evidence--claim linkage, without directly mirroring rubric definitions, and serve as a shared intermediate representation between essays and rubric labels. During training, we use them as intermediate supervision, either by providing them as additional inputs or by jointly predicting them with the final rubric label (see Section~\ref{sec:traits}). Table~\ref{tab:intro_traits} summarizes the traits used in our framework along with their brief descriptions.

In addition to rubric-agnostic traits as one component of our framework, we study a second component: how does the level of supervision available on \emph{essays} (and seen rubrics) during training affects generalization. We consider three settings: \textbf{No Labels}, the hardest setting, where target essays are not observed during training; \textbf{Pseudo Labels}, where target essays are added using automatically generated labels under seen rubrics; and \textbf{Gold Labels}, where target essays are added using human-annotated labels under seen rubrics (see Section~\ref{sec:supervision}). In all cases, the target rubric remains unseen during training.

We jointly examine these two components and find that both are important for cross-rubric generalization, but in different ways. Traits are especially helpful in the hardest \emph{No Labels} setting, where the model has no access to target essays during training, improving macro F1 by 5.0\% compared to a baseline without traits. Performance improves further as more target-essay supervision becomes available. Under the strongest supervision setting, our best model outperforms the proprietary prompting-only GPT-5-mini by 2.1\% in macro F1 and trails GPT-5 by only 1.9\%. Our contributions are as follows: (1) We study \emph{cross-rubric generalization} in automated essay scoring, grounded in critical thinking essay scoring across six rubrics. (2) We present a systematic study of two key components: rubric-agnostic \emph{intermediate} trait representations and target-essay supervision from seen rubrics during training. (3) We show that these two components improve generalization in complementary ways: traits help in the hardest setting, where no target-essay labels are available under seen rubrics, while the strongest target-essay supervision enables a fine-tuned open-source model to approach proprietary prompting-based models.

\begin{table}[]
\caption{Rubric-agnostic analytic traits used in our framework for critical thinking essay scoring.}
\centering
\small
\renewcommand{\arraystretch}{1.1}
\begin{tabularx}{\linewidth}{>{\raggedright\arraybackslash}X}
\toprule
\textbf{Trait} \\
\midrule
\textbf{Claim Explicitness}: how clearly the essay states its main point or position \\

\textbf{Structural Coherence}: how well the ideas are organized and connected across the essay \\

\textbf{Supporting Evidence}: how much the essay uses examples, facts, or details to support its points \\

\textbf{Evidence--Claim Linkage}: how clearly the essay explains why the evidence supports its main point \\

\textbf{Alternative Perspectives}: how much the essay considers and responds to other viewpoints on the issue \\

\textbf{Analytical Depth}: how much the essay goes beyond description to compare, evaluate, or reason ideas \\
\bottomrule
\end{tabularx}
\label{tab:intro_traits}
\end{table}


\section{Related Work}

\noindent\textbf{Generalization in Automated Essay Scoring}. 
A large body of work in automated essay scoring (AES) studies generalization beyond the training distribution, with the dominant focus on \emph{cross-prompt} transfer. In this setting, models are evaluated on essays from unseen prompts, with the primary source of shift being the essay prompt rather than the target rubric. Prior work has explored this problem through cross-prompt trait scoring \citep{Ridley_He_Dai_Huang_Chen_2021}, grammar-aware modeling for prompt transfer \citep{do-etal-2025-towards}, rubric-derived assessment questions and features \citep{eltanbouly-etal-2025-trates}, and human--AI collaborative scoring pipelines \citep{xiao2025human}. In contrast, we study generalization across rubrics, training on one set of rubrics and evaluating on a previously unseen set. Closest to our work, \citep{peczuh2025toward} study generalization to previously unseen critical-thinking rubrics by conditioning on rubric text, but in a setting that primarily isolates rubric shift rather than requiring simultaneous generalization across both essays and rubrics. Our work extends this line of research in two ways. First, we cast the problem more broadly as \emph{cross-rubric generalization} and study it under multiple generalization conditions, including both joint essay-and-rubric shift and settings that isolate rubric shift. Second, we systematically analyze two components for improving generalization to unseen rubrics: rubric-agnostic intermediate trait representations and target-essay supervision during training.


\noindent\textbf{Rubric-Grounded Structured Scoring}. 
Recent LLM-based essay scoring methods increasingly use rubrics to construct intermediate structure rather than relying only on direct end-to-end grading. For example, prior work has proposed trait-focused conversational decomposition \citep{lee-etal-2024-unleashing}, rubric-derived assessment questions and features \citep{eltanbouly-etal-2025-trates}, trait-specific rationales \citep{chu-etal-2025-rationale}, multi-agent extraction of rubric-relevant components \citep{Wang_Ding_Wu_Sun_Liu_Zhai_2026}, and structured human--AI scoring workflows \citep{xiao2025human}. Related work outside AES also points to the value of portable structured signals and interpretable latent factors \citep{al-khatib-etal-2020-exploiting, lan-etal-2025-autoqual}. Our work is closely related to this line of research, but differs in both setting and scope. We study generalization to \emph{previously unseen rubrics}, rather than settings where evaluation remains within seen rubrics or where the primary shift is across prompts, and we systematically analyze two components that may support transfer in this setting: rubric-agnostic trait annotations and target-essay supervision during training.

\section{Task Setup}

\subsection{Task Definition}
\label{sec:task_definition}

Rubric-based assessment requires models to map essays to discrete proficiency levels defined by natural-language rubrics, and poses a key challenge of generalizing to evaluation criteria not seen during training. Let $\mathcal{E}$ denote a set of essays and $\mathcal{R}=\{r_1,\dots,r_K\}$ denote a set of rubrics, where each rubric defines an evaluation criterion in natural language. Each rubric $r\in\mathcal{R}$ defines a labeling function
\[
f_r:\mathcal{E}\rightarrow\mathcal{Y}_r,
\]
where $\mathcal{Y}_r$ is the associated discrete proficiency-level scale, i.e., a set of ordinary scores.
We partition essays and rubrics into disjoint subsets $\mathcal{E}_{\text{train}}, \mathcal{E}_{\text{test}}$ and $\mathcal{R}_{\text{train}}, \mathcal{R}_{\text{test}}$, respectively, with $\mathcal{E}_{\text{train}} \cup \mathcal{E}_{\text{test}}=\mathcal{E}$ and $\mathcal{R}_{\text{train}} \cup \mathcal{R}_{\text{test}}=\mathcal{R}$.
We denote the labeled training data, provided by human annotation, as
\begin{align*}
\mathcal{D}_{\text{train}}&=\{(e,r,y)\mid e\in\mathcal{E}_{\text{train}},\, r\in\mathcal{R}_{\text{train}},\, y=f_r(e)\}.
\end{align*}
A model $M$ is trained to approximate $f_r$ for rubrics in $\mathcal{R}_{\text{train}}$ using $\mathcal{D}_{\text{train}}$ as the base training set. Depending on the supervision setting, this base set may be augmented with essays from $\mathcal{E}_{\text{test}}$ and/or rubrics from $\mathcal{R}_{\text{train}}$ (Section~\ref{sec:supervision}). At test time, the model is evaluated on
\begin{align*}
\mathcal{D}_{\text{test}}&=\{(e,r',y)\mid e\in\mathcal{E}_{\text{test}},\, r'\in\mathcal{R}_{\text{test}},\, y=f_{r'}(e)\},
\end{align*}
without access to labeled data from $\mathcal{R}_{\text{test}}$ during training. The model is provided with the textual description of the target rubric $r'\in\mathcal{R}_{\text{test}}$.
The objective is to evaluate whether $M$ can approximate $f_{r'}$ for unseen rubrics on essays in $\mathcal{E}_{\text{test}}$, requiring generalization to new rubric-defined label functions. The hardest supervision setting additionally involves generalization to unseen essays.

\subsection{Dataset}

We instantiate our setup using the dataset of \citep{peczuh2025toward}, which contains 500 student essays from the public portion of the PERSUADE 2.0 corpus. These essays are argumentative writing samples from 6th--12th grade U.S. students, 
covering civic, scientific, and social topics. The dataset defines six rubrics, organized under three broader skills: \textit{Information Analysis}, \textit{Argument Generation}, and \textit{Logical Reasoning}, with two rubrics per skill. Each rubric is specified in natural language through a name and definition and a rubric description, which together form the \textit{rubric specification}, and assigns essays a 5-point rubric label (0--4: \textit{Not Applicable}, \textit{Below Emerging}, \textit{Emerging}, \textit{Expanding}, \textit{Exemplifying}). Because each essay is annotated under all six rubrics, the dataset allows us to decouple essays from rubrics and study cross-rubric generalization. Appendix Table~\ref{table:dataset-subskills} lists the rubric names and definitions.

\section{Methods}
\label{sec:methods}

In this section, we detail our approach to cross-rubric generalization within an LLM fine-tuning framework for rubric-based essay scoring. We first present the core rubric assessment setup, then introduce trait-based representations as rubric-agnostic intermediate structure, and finally define supervision settings that vary the availability of essay--rubric labels during training while preventing leakage from unseen rubrics.

\subsection{LLM fine-tuning framework for rubric-based essay scoring}

We build on the LLM-based framework for rubric-based essay scoring introduced by \citep{peczuh2025toward}, which formulates rubric-based evaluation as a conditional text generation task. Specifically, we fine-tune a pretrained open-source LLM ($M$) (Llama 3.1 8B Instruct \citep{grattafiori2024llama}) using standard cross-entropy loss to predict rubric labels for essay--rubric pairs. For each essay--rubric pair, the model takes as input the essay text together with the full rubric specification and is trained to generate a natural-language justification followed by the corresponding rubric label, yielding a rationale-then-decision output format \citep{wei2022chain}. The justification is a short explanation of why the essay satisfies the gold rubric label, and is generated by a separate LLM prior to fine-tuning. Appendix~\ref{sec:res_desc_and_just} provides ablations on rubric specification and justifications. Training uses the base training set $\mathcal{D}_{\text{train}}$, while evaluation is conducted on $\mathcal{D}_{\text{test}}$. Across all settings, rubrics in $\mathcal{R}_{\text{test}}$ remain unseen during training; the availability of essays from $\mathcal{E}_{\text{test}}$ under rubrics in $\mathcal{R}_{\text{train}}$ varies by supervision setting (Section~\ref{sec:supervision}).


\subsection{Trait-Based Representations}
\label{sec:traits}

To improve generalization to previously unseen rubrics, we introduce \textit{rubric-agnostic analytic traits} as an intermediate representation between essays and rubric-specific labels. Prior work in AES shows that structured intermediate features can improve transfer by capturing reusable aspects of writing quality \citep{eltanbouly-etal-2025-trates}. We extend this idea to the cross-rubric setting by grounding these traits in argumentation and critical thinking theory, which characterizes writing quality in terms of lower-level components such as claims, evidence, reasoning, and perspective-taking \citep{toulmin2003uses, kuhn1991skills, facione1990critical}. 

Concretely, we decompose rubric-based scoring as:
$e \;\rightarrow\; h(e) \;\rightarrow\; f_r(e),$
where $h(e)$ captures rubric-independent argumentative properties. This factorization encourages the model to learn transferable reasoning representations that can be flexibly mapped to different rubric-specific criteria.

\noindent\textbf{Trait Schema}. 
We define a fixed set of six analytic traits derived from established frameworks in argumentation theory and critical thinking. These traits are (i) lower-level criteria that underlie rubrics, (ii) shared across all rubrics, and (iii) defined prior to experimentation and held fixed across train/test splits. The six traits are:
\textit{Claim Explicitness}, 
\textit{Structural Coherence}, 
\textit{Presence of Supporting Evidence}, 
\textit{Evidence--Claim Linkage}, 
\textit{Consideration of Alternative Perspectives}, and 
\textit{Analytical Depth}. 
Each trait is associated with a three-level categorical value scale (e.g., absent/partial/strong), capturing increasing proficiency. These traits are designed to capture fundamental argumentative properties that are shared across all rubric definitions. Table~\ref{table:traits} in the Appendix summarizes each trait, along with its annotation question and value scale.

\noindent\textbf{Trait Annotation}. 
For each essay $e \in \mathcal{E}$, we use an LLM to generate trait annotations consisting of (i) a categorical \textit{value} for each trait and (ii) a short natural language \textit{observation} explaining the label assignment. These annotations define a rubric-agnostic representation $h(e)$ shared across all rubrics. Trait annotations are automatically generated (silver labels) for all essays in $\mathcal{E}$, and do not use labeled data from $\mathcal{R}_{\text{test}}$, ensuring compatibility with the cross-rubric setting. We acknowledge that these annotations are not validated through human annotation and treat them as 
intermediate structure that supports training, leaving human validation to future work.

\noindent\textbf{Trait-Augmented Fine-Tuning}. 
We incorporate trait representations into model fine-tuning in four configurations, depending on whether the model uses only trait \textit{values} or both trait \textit{values} and natural-language \textit{observations}, and whether these traits are provided as inputs or predicted as outputs. In \textit{Input: Trait Values}, we append only the categorical trait values to the input alongside the essay and rubric. In \textit{Input: Trait Values + Observations}, we append both the trait values and their corresponding natural-language observations. In \textit{Output: Trait Values}, the model is trained to jointly predict the categorical trait values together with the final rubric label. In \textit{Output: Trait Values + Observations}, the model is trained to jointly generate the trait values, the accompanying observations, and the final rubric label. These variants let us test whether traits are more useful as input scaffolds or as auxiliary prediction targets for generalization to $(e, r')$ pairs with $e \in \mathcal{E}_{\text{test}}$ and $r' \in \mathcal{R}_{\text{test}}$.

\subsection{Supervision Settings}
\label{sec:supervision}

We consider three supervision settings that differ in whether essays from $\mathcal{E}_{\text{test}}$ are observed during training under rubrics in $\mathcal{R}_{\text{train}}$, while ensuring that rubrics in $\mathcal{R}_{\text{test}}$ are never used for supervision:
\begin{itemize}[itemsep=0em]
    \item In the \textbf{No Labels} setting, the model $M$ is trained on $\mathcal{D}_{\text{train}} = \{(e, r, y) \mid e \in \mathcal{E}_{\text{train}},\; r \in \mathcal{R}_{\text{train}}\}$ and evaluated on $\mathcal{D}_{\text{test}} = \{(e, r', y) \mid e \in \mathcal{E}_{\text{test}},\; r' \in \mathcal{R}_{\text{test}}\}$. This setting represents the hardest generalization scenario where we evaluate the model's ability to generalize to both unseen essays and unseen rubrics.
    \item In the \textbf{Pseudo Labels} setting, we augment training data using self-training. We first train a model $M$ in the No Labels setting. We then use $M$ to generate pseudo labels for $(e, r)$ where $e \in \mathcal{E}_{\text{test}}$ and $r \in \mathcal{R}_{\text{train}}$. These pseudo-labeled examples, corresponding to essays from $\mathcal{E}_{\text{test}}$, are combined with $\mathcal{D}_{\text{train}}$ to train a new model $M'$. The resulting model $M'$ is evaluated on $(e, r')$ with $e \in \mathcal{E}_{\text{test}}$ and $r' \in \mathcal{R}_{\text{test}}$. This setting provides the model with training signals from $\mathcal{E}_{\text{test}}$ without using human-annotated labels.
    \item In the \textbf{Gold Labels} setting, essays from $\mathcal{E}_{\text{test}}$ are available during training, but only under rubrics in $\mathcal{R}_{\text{train}}$. Specifically, we train $M$ on $\{(e, r, y) \mid e \in \mathcal{E}_{\text{train}} \cup \mathcal{E}_{\text{test}},\; r \in \mathcal{R}_{\text{train}}\}$, using gold (human-annotated) labels for all essay--rubric pairs under $\mathcal{R}_{\text{train}}$. The model is then evaluated on $(e, r')$ with $e \in \mathcal{E}_{\text{test}}$ and $r' \in \mathcal{R}_{\text{test}}$. This setting is closest to \citep{peczuh2025toward}, but differs in that evaluation is performed only on $\mathcal{E}_{\text{test}}$, not on the full essay set $\mathcal{E}$. It therefore isolates generalization across rubrics while preserving a held-out essay split at evaluation time. Importantly, in both Pseudo and Gold Labels settings, essays from $\mathcal{E}_{\text{test}}$ are observed during training only under $\mathcal{R}_{\text{train}}$, and no information from $\mathcal{R}_{\text{test}}$ is used for supervision.
\end{itemize}

\section{Experimental Setup}

We detail the experimental setup for evaluating cross-rubric generalization. We first define the structured design space explored in our experiments, together with the baselines, and then the dataset splitting strategy, and evaluation metrics.

\subsection{Design Space, and Baselines}

We define a unified experimental framework consisting of (i) a design space of modeling choices based on the methods detailed above, (ii) a baseline configuration based on our method, and (iii) prompting-based baselines. We detail each of these components below.


\noindent\textbf{Design space}. 
Our experiments study cross-rubric generalization within the LLM-based framework for rubric-based essay scoring detailed in Section~\ref{sec:methods}. We study a combinatorial design space over two complementary components: (i) trait-based intermediate representations (Section~\ref{sec:traits}), and (ii) target-essay supervision settings (Section~\ref{sec:supervision}). This design enables us to evaluate both the individual and combined effects of these two components.

\noindent\textbf{Fine-tuning Baseline}. Our primary baseline is the configuration without traits or target-essay labels \textit{(No Labels, No Traits)}. It reflects the standard cross-rubric generalization scenario: the model is trained on essays and rubrics available during training and evaluated on unseen essay--rubric pairs. All other configurations extend this baseline by adding trait representations, target-essay supervision, or both.

\noindent\textbf{Prompting-based Baselines}.
We compare against three prompting-based baselines. Two are proprietary LLMs (GPT-5-mini and GPT-5), following \citep{peczuh2025toward} for rubric label generation. In the zero-shot setting, these models receive the essay and the target rubric specification at inference time ($r' \in \mathcal{R}_{\text{test}}$), without rubric-specific training, and output both a proficiency label and a concise justification explaining the prediction. Their ability to reason directly over unseen rubrics makes them strong baselines for cross-rubric generalization. The third baseline is Llama 3.1 8B Instruct \citep{grattafiori2024llama} used in a zero-shot prompting setting (without fine-tuning), receiving the same essay and rubric specification and outputting only a proficiency label. This baseline allows us to isolate the contribution of fine-tuning by comparing zero-shot prompting and fine-tuned variants of the same underlying model. In contrast to all prompting-based baselines, our fine-tuning approach learns transferable representations over seen rubrics. See Appendix~\ref{sec:implementation} for the implementation details of our experiments.


\subsection{Dataset Splitting}

We partition the set of essays $\mathcal{E}$ into disjoint subsets $\mathcal{E}_{\text{train}}$ and $\mathcal{E}_{\text{test}}$ using an 80/20 split. From $\mathcal{E}_{\text{train}}$, we further reserve 10\% as a validation set, resulting in an effective 70/10/20 split for train/validation/test essays.
We also partition the set of rubrics $\mathcal{R}$ according to their organization into three skills. For each of the three possible choices of held-out skill, we designate the two rubrics under that skill as $\mathcal{R}_{\text{test}}$ and use the remaining four rubrics from the other two skills as $\mathcal{R}_{\text{train}}$. This rubric partitioning ensures that all test rubrics are unseen during training.
The full splitting procedure yields three rubric partitions, each with its own $\mathcal{R}_{\text{train}}$--$\mathcal{R}_{\text{test}}$ split. We train and evaluate one model for each rubric partition, concatenate predictions across the three held-out test sets, and compute evaluation metrics on the combined predictions.

\subsection{Evaluation Metrics}

Following \citep{peczuh2025toward}, we report Accuracy, Macro F1 (denoted as F1), and Krippendorff's $\alpha$ (denoted as $\alpha$). Accuracy measures the proportion of essay--rubric pairs for which the predicted label exactly matches the gold label. Macro F1 averages per-label F1 scores, giving equal weight to all labels, better reflecting performance on less frequent ones; we treat it as our primary metric. Krippendorff's $\alpha$ measures agreement with gold labels while accounting for both chance agreement and the ordinal structure of the proficiency scale, making it well suited for this graded evaluation setting. To assess statistical significance, we use paired bootstrap resampling \citep{efron1994introduction} with 2{,}000 resamples on macro F1; results are considered significant at $p$<0.05. All comparisons are made against the \textit{(No Labels, No Traits)} baseline.


\begin{table*}[!t]
\caption{Comparing different trait configurations and supervision settings proposed in our paper. \textit{No Traits} under \textit{No Labels} is our fine-tuning baseline, while Zero-Shot prompted models serve as prompting baselines. Best numbers in each column are \textbf{bolded}.}
\centering
\small
\setlength{\tabcolsep}{5pt}
\renewcommand{\arraystretch}{1.1}
\begin{tabular}{lccc|ccc|ccc}
\toprule
& \multicolumn{3}{c|}{\textbf{No Labels}} 
& \multicolumn{3}{c|}{\textbf{Pseudo Labels}} 
& \multicolumn{3}{c}{\textbf{Gold Labels}} \\
\cmidrule(lr){2-4} \cmidrule(lr){5-7} \cmidrule(lr){8-10}
\textbf{Trait Configuration / Model} & \textbf{Acc} & \textbf{F1} & $\pmb{\alpha}$ & \textbf{Acc} & \textbf{F1} & $\pmb{\alpha}$ & \textbf{Acc} & \textbf{F1} & $\pmb{\alpha}$ \\
\midrule
No Traits                             & 0.399 & 0.353 & 0.346 & 0.412 & 0.376 & \textbf{0.368} & 0.470 & \textbf{0.431}$^{*}$ & \textbf{0.412} \\
Input: Trait Values                   & 0.406 & 0.367 & 0.363 & \textbf{0.446} & \textbf{0.400}$^{\dagger}$ & 0.360 & 0.473 & \textbf{0.433}$^{**}$ & 0.378 \\
Input: Trait Values and Observations  & 0.382 & 0.341 & 0.291 & 0.417 & 0.369 & 0.366 & 0.451 & 0.405$^{\dagger}$ & 0.293 \\
Output: Trait Values                  & \textbf{0.436} & \textbf{0.403}$^{*}$ & \textbf{0.363} & 0.416 & 0.383 & 0.306 & \textbf{0.495} & 0.415$^{*}$ & 0.369 \\
Output: Trait Values and Observations & 0.387 & 0.329 & 0.195 & 0.424 & 0.360 & 0.251 & 0.444 & 0.373 & 0.392 \\
\midrule
Llama 3.1 8B Instruct (Zero-shot)     & 0.372 & 0.222 & 0.251 & \multicolumn{3}{c|}{---} & \multicolumn{3}{c}{---} \\
GPT-5-mini (Zero-shot)                & 0.492 & 0.410$^{\dagger}$ & 0.388 & \multicolumn{3}{c|}{---} & \multicolumn{3}{c}{---} \\
GPT-5 (Zero-shot)                     & \textbf{0.532} & \textbf{0.450}$^{*}$ & \textbf{0.476} & \multicolumn{3}{c|}{---} & \multicolumn{3}{c}{---} \\
\bottomrule
\multicolumn{10}{l}{\footnotesize Markers on F1 denote paired bootstrap significance vs.\ \textit{(No Labels, No Traits)} baseline: $^{\dagger}$p < 0.05, $^{*}$p < 0.01, $^{**}$p < 0.001.}
\end{tabular}
\label{tab:design_choice}
\end{table*}

\section{Results}

We present results for cross-rubric generalization. We first analyze the impact of key design choices, including trait-based representations and supervision settings, and then compare our fine-tuned models against the prompting baselines. Table~\ref{tab:design_choice} summarizes results across trait-based representations and supervision settings.

\subsection{Impact of Design Choices on Cross-Rubric Generalization}

We observe three main trends: (i) incorporating trait-based representations improves performance in the absence of direct supervision (\textit{No Labels}), (ii) the benefit of traits diminishes as supervision becomes available (\textit{Pseudo Labels} and \textit{Gold Labels}), and (iii) increasing supervision consistently improves performance, with \textit{Gold Labels} providing the strongest gains and \textit{Pseudo Labels} remaining competitive.

\paragraph{Traits improve performance in the No Labels setting:}
We observe that in the \textit{No Labels} setting, incorporating traits improves performance compared to the \textit{No Traits} baseline. In particular, using \textit{Output: Trait Values} improves accuracy from 0.399 to 0.436 (+3.7\%), F1 from 0.353 to 0.403 (+5.0\%, $p$=0.006), and $\alpha$ from 0.346 to 0.363 (+1.7\%). We further observe that predicting trait values as outputs performs better than providing them as inputs, since generating trait-level abstractions encourages the model to learn a structured mapping from text to rubric-relevant features. Incorporating trait observations does not improve performance, since longer textual explanations introduce variability that makes it harder to extract consistent signals. This result indicates that trait values provide useful intermediate supervision, enabling better generalization in the absence of direct rubric-specific labels during training. See Section~\ref{sec:case_studies} and Table~\ref{tab:trait_case_studies} in the Appendix for qualitative examples where \textit{Output: Trait Values} predicts the correct rubric label while the baseline \textit{No Traits} does not.

\paragraph{Traits provide limited gains in the Pseudo Labels and Gold Labels settings:}
In contrast, under the \textit{Pseudo Labels} and \textit{Gold Labels} settings, trait-based configurations do not consistently outperform their corresponding \textit{No Traits} baselines across all metrics. In the \textit{Pseudo Labels} setting, \textit{Input: Trait Values} improves accuracy from 0.412 to 0.446 (+3.4\%) and F1 from 0.376 to 0.400 (+2.4\%), while achieving a comparable $\alpha$ (0.368 vs.\ 0.360). In the \textit{Gold Labels} setting, \textit{Output: Trait Values} improves accuracy from 0.470 to 0.495 (+2.5\%), but does not improve F1 (0.431 to 0.415) or $\alpha$ (0.412 to 0.369). These results suggest that when rubric-label supervision on target essays is available, the model can better learn rubric-agnostic essay representations under the training rubrics, thereby reducing the benefit of intermediate trait-based representations.

\paragraph{Increasing supervision improves performance, with Gold Labels providing the strongest gains:}
We observe that increasing supervision from \textit{No Labels} to \textit{Pseudo Labels} to \textit{Gold Labels} leads to consistent improvements in performance in the \textit{No Traits} setting. Moving from \textit{No Labels} to \textit{Pseudo Labels} improves accuracy from 0.399 to 0.412 (+1.3\%), F1 from 0.353 to 0.376 (+2.3\%, $p$=0.109), and $\alpha$ from 0.346 to 0.368 (+2.2\%). Moving from \textit{No Labels} to \textit{Gold Labels} improves accuracy from 0.399 to 0.470 (+7.1\%), F1 from 0.353 to 0.431 (+7.8\%, $p$<0.01), and $\alpha$ from 0.346 to 0.412 (+6.6\%). This result indicates that higher-quality supervision provides stronger and more reliable signals for generalization across rubrics, with the \textit{Gold Labels} gain reaching statistical significance ($p$<0.01) while the \textit{Pseudo Labels} gain, though consistent, remains modest, making \textit{Pseudo Labels} a practical alternative when gold supervision is unavailable.

\subsection{Comparison with Prompting-based Baselines}

We observe two main trends: (i) fine-tuning substantially outperforms zero-shot prompting of the same underlying open-source model, confirming that the gains are attributable to fine-tuning rather than model scale alone, and (ii) fine-tuned open-source LLMs are competitive with proprietary LLMs, achieving performance comparable to smaller models and approaching larger ones.

\paragraph{Fine-tuning substantially outperforms zero-shot prompting of the same model:}
We compare Llama 3.1 8B Instruct under zero-shot prompting against the fine-tuning baseline (No Labels, No Traits). Moving from zero-shot prompting to the fine-tuning baseline most notably improves F1 from 0.222 to 0.353 (+13.1\%, $p$<0.001), with additional gains in accuracy from 0.372 to 0.399 (+2.7\%) and $\alpha$ from 0.251 to 0.346 (+9.5\%). We observe that accuracy remains similar across both settings, but F1 and $\alpha$ are substantially lower under zero-shot prompting, since without exposure to the true label distribution, the model concentrates predictions on a few frequent proficiency levels, collapsing recall on minority labels and reducing ordinal alignment with human judgments. This result indicates that fine-tuning over seen rubrics provides the calibration needed to use the full label space, and that these gains cannot be achieved through zero-shot in-context reasoning alone.

\paragraph{Fine-tuned open-source models achieve competitive performance with prompting proprietary models:}
We observe that our best fine-tuned model (Gold Labels) achieves performance comparable to GPT-5-mini. Compared to GPT-5-mini, our model achieves slightly higher F1 from 0.410 to 0.431 (+2.1\%) and $\alpha$ from 0.388 to 0.412 (+2.4\%), while accuracy changes from 0.492 to 0.470. Compared to GPT-5, our model achieves comparable F1, with values changing from 0.450 to 0.431, but lower accuracy from 0.532 to 0.470 and lower $\alpha$ from 0.476 to 0.412. This result indicates that fine-tuning can approach the performance of lightweight proprietary models and remain competitive with larger ones. This result is particularly important in practice, since fine-tuned models provide advantages in data privacy, deployment control, and reduced inference costs compared to accessing proprietary LLMs through API calls.

\section{Conclusion and Future Work}

In this work, we introduced \emph{cross-rubric generalization} as a new setting for automated essay scoring, where models must generalize to previously unseen rubrics. Within an LLM fine-tuning framework for rubric-based essay scoring, we studied rubric-agnostic trait annotations and target-essay supervision during training. Our results show that traits are especially useful in the hardest setting, when no target-essay supervision is available, improving macro F1 by 5.0\%, while increasing target-essay supervision yields further gains. Our best fine-tuned open-source Llama-based model outperforms the proprietary prompting-only GPT-5-mini by 2.1\% macro F1 and trails GPT-5 by only 1.9\%, suggesting that fine-tuned open-source models can provide a scalable, cost-effective, and privacy-preserving alternative for rubric-based assessment as scoring criteria evolve.

Future work can be done in the following directions. First, our findings are based on a single critical thinking dataset with six rubrics, and future work could examine the extent to which these results generalize to other datasets, rubric designs, and writing domains. Second, gains from pseudo labels suggest that stronger low-supervision learning, such as improved self-training, may further reduce noise in pseudo-labeled target essays. Third, our findings motivate learning more robust rubric-agnostic essay representations through contrastive learning or prototypical classification approaches that better capture shared argumentative structure across rubrics. Fourth, trait predictions can not only offer intermediate supervision, but also reward signals for training objectives that directly optimize cross-rubric generalization, similar to recent reinforcement-learning-based work on multi-trait essay scoring \citep{do-etal-2024-autoregressive-multi}.


\newpage
\section*{Declaration on Generative AI}
During the preparation of this work, the authors used ChatGPT (GPT-5.4) for paraphrasing and rewording text. After using this tool, the authors reviewed and edited the content as needed and take full responsibility for the publication's content.

\bibliography{sample-ceur}

\appendix

\section{Appendix}
\label{sec:appendix}

\subsection{Rubric Details}
Table~\ref{table:dataset-subskills} summarizes the rubric details in the critical thinking dataset introduced by \citep{peczuh2025toward}. Following our terminology in this paper, we refer to these dataset-defined subskills as rubrics.

\subsection{Trait Schema}
Table~\ref{table:traits} shows the trait schema for the rubric-agnostic analytic traits used in our framework, including the annotation question and value scale associated with each trait. These traits capture lower-level argumentative properties that may generalize across rubrics.

\subsection{Prompts}

Figure~\ref{fig:trait-prompt} shows the prompt used to generate trait annotations for each essay. The \texttt{\{TRAIT\_SCHEMA\}}, including annotation questions and value scales, is provided in Table~\ref{table:traits}. The placeholder \texttt{\{ESSAY\_TEXT\}} represents the essay being annotated.

Figure~\ref{fig:justification-prompt} shows the prompt used to generate justifications for gold rubric labels. The placeholders \texttt{\{PROFICIENCY\_LEVEL\_DEFINITIONS\}}, \texttt{\{subskill\}}, \texttt{\{definition\}}, and \texttt{\{RUBRIC\_DEFINITION\}} correspond to the general proficiency-level definitions, subskill (rubric) names and definitions, and the corresponding rubric descriptions from \citep{peczuh2025toward}. The placeholders \texttt{\{essay\}} and \texttt{\{gt\_label\}} denote the essay being evaluated and its gold label for the given subskill, respectively.

Figure~\ref{fig:prompting-rubric-label-generation} shows the prompt used for prompting-based baseline for rubric label generation. The placeholders \texttt{\{PROFICIENCY\_LEVEL\_DEFINITIONS\}}, \texttt{\{subskill\}}, \texttt{\{definition\}}, and \texttt{\{RUBRIC\_DESCRIPTION\}} correspond to the general proficiency-level definitions, subskill (rubric) names and definitions, and the corresponding rubric descriptions from \citep{peczuh2025toward}. The placeholder \texttt{\{essay\}} denotes the essay to be rated.

\begin{table*}[t]
\centering
\renewcommand{\arraystretch}{1.15}
\begin{tabularx}{\linewidth}{p{4.0cm} p{4.5cm} X}
\toprule
\textbf{Rubric (Subskill) Name} & \textbf{Rubric (Subskill) Definition} & \textbf{Illustrative Example} \\
\midrule

\multicolumn{3}{l}{\textbf{2: Information Analysis}} \\
2.1 Synthesizing Multiple Sources
& Effectively synthesizes multiple pieces of information
& Summarizes information from multiple sources (with citations) but does not integrate information \\

2.2 Evaluating Evidence Strength
& Evaluates the strength and relevance of evidence used to form a conclusion
& Presents evidence and links evidence to a specific conclusion, but does not evaluate the relevance or strength of the evidence for the argument generated \\

\addlinespace[0.3em]
\multicolumn{3}{l}{\textbf{3: Argument Generation}} \\
3.1 Using Counterarguments
& Effectively addresses counterarguments
& Acknowledges specific opposing viewpoint(s), counterargument(s), or qualifier(s) \\

3.2 Using Facts and Opinions
& Relies on data and/or facts over opinions
& Uses facts and opinions about equally to support claim(s) and/or arguments \\

\addlinespace[0.3em]
\multicolumn{3}{l}{\textbf{4: Logical Reasoning}} \\
4.1 Drawing Conclusions
& Draws specific conclusions
& Draws a specific conclusion to analyze simple and straightforward relationships/argumentations \\

4.2 Using Logical Fallacies
& Recognizes and avoids logical fallacies
& Uses logical fallacies and evidence about equally when generating arguments \\

\bottomrule
\end{tabularx}
\caption{Subskill details in the critical thinking dataset, adapted from \citep{peczuh2025toward}. We retain the original table content from their paper while changing the column headers to match the terminology used in this paper.}
\label{table:dataset-subskills}
\end{table*}

\begin{figure*}[htbp]
\centering
\begin{tcolorbox}[
    colback=gray!5!white,
    colframe=black,
    title=Prompt for Trait Annotation Generation
]

\textbf{System Prompt}

You are an expert writing evaluator analyzing the argumentative quality of an essay.

\vspace{0.5em}
Your task is to assign analytic trait labels that describe the essay's reasoning and argument structure.
For each trait below, choose exactly one label from the allowed list.

\vspace{0.5em}
For each trait, first write a brief observation from the essay (a direct quote or paraphrase that supports your judgment), then assign the label.
Return ONLY a valid JSON object (parseable by \texttt{json.loads}).
Do not include any text outside the JSON object.
If a trait is genuinely ambiguous, choose the label that best matches the essay's dominant tendency.

\vspace{0.5em}
\textbf{Traits}

\texttt{\{TRAIT\_SCHEMA\}} is inserted here. Refer to Table~\ref{table:traits} for the trait schema, which contains the trait names, annotation questions, and value scales.

\vspace{0.5em}
\textbf{Output format (exact keys required):}

\begin{quote}
\ttfamily
\{ \\
"claim\_explicitness": \{"observation": "...", "label": "..."\}, \\
... similar structure for all other traits \\
\}
\end{quote}

\textbf{User Prompt}

Essay:

\begin{quote}
\ttfamily
<<< \\
\{ESSAY\_TEXT\} \\
>>>
\end{quote}

\end{tcolorbox}
\caption{Prompt used to generate trait annotations for each essay. The \texttt{\{TRAIT\_SCHEMA\}} is provided in Table~\ref{table:traits}. The placeholder \texttt{\{ESSAY\_TEXT\}} represents the essay being annotated.}
\label{fig:trait-prompt}
\end{figure*}

\begin{figure*}[htbp]
\centering
\begin{tcolorbox}[
    colback=gray!5!white,
    colframe=black,
    title=Prompt for Justification Generation
]

\textbf{System Prompt}

You are a calibrated educational scorer designed to provide evidence-based justifications for a gold proficiency label on a student essay. Given an essay, a critical thinking subskill rubric, and a gold proficiency level (0--4), your task is to generate a concise justification that explains why the essay merits that specific level according to the rubric criteria. Maintain objectivity, ensure your justification strictly aligns with the rubric definitions, and base your explanation only on information presented in the essay. Do not make assumptions, incorporate external knowledge, or question the assigned level---your role is solely to justify the given classification with evidence from the text.

\vspace{1em}
\textbf{User Prompt}

Your job is to evaluate why a gold label was provided for a specific essay and subskill.

\vspace{0.5em}
\texttt{\{PROFICIENCY\_LEVEL\_DEFINITIONS\}} is inserted here. Refer to \citep{peczuh2025toward} for the general definitions of the proficiency levels used in the dataset (\textit{Not Applicable}, \textit{Below Emerging}, \textit{Emerging}, \textit{Expanding}, and \textit{Exemplifying}).

\vspace{0.5em}
You will be given a subskill and its definition, followed by the rubric for that subskill and the gold label for the essay. Your task is to explain why this gold label was assigned based on the rubric and provide a 1--2 sentence justification for that choice.

\vspace{0.5em}
\textbf{Output format:}
\begin{itemize}[noitemsep, topsep=0pt]
    \item First line: \texttt{Label - <number>: <label name>}
    \item Second line: \texttt{Justification - <your justification>}
\end{itemize}

\vspace{0.5em}
\textbf{Input}

\vspace{0.5em}
\textbf{Subskill Information}\\
\texttt{Subskill - \{subskill\}}\\
\texttt{Definition - \{definition\}}

\vspace{0.5em}
\texttt{\{RUBRIC\_DESCRIPTION\}} is inserted here. Refer to \citep{peczuh2025toward} for the rubric descriptions associated with the given subskill.

\vspace{0.5em}
\textbf{Rated Essay}\\
\texttt{\{essay\}}

\vspace{0.5em}
\textbf{Gold Label for Particular Subskill}\\
\texttt{\{gt\_label\}}

\vspace{0.5em}
\textbf{Output of Justification}\\
Strictly adhere to the prescribed output format above. Do not enter any additional words, commentary, or preambles; only the two required lines.

\end{tcolorbox}
\caption{Prompt used to generate justifications for gold rubric labels. The general proficiency-level definitions, subskill names and definitions, and the corresponding rubric descriptions are taken from \citep{peczuh2025toward}.}
\label{fig:justification-prompt}
\end{figure*}

\begin{figure*}[htbp]
\centering
\begin{tcolorbox}[
    colback=gray!5!white,
    colframe=black,
    title=Prompt for Prompting-based Rubric Label Generation
]

\textbf{System Prompt}

You are a calibrated educational scorer designed to classify student essays on fine-grained critical thinking subskills using a structured rubric. For each essay, assign one of five proficiency levels (0--4) and provide a concise, evidence-based justification strictly aligned with the rubric. Maintain objectivity, ensure consistency, and base your evaluation only on information presented in the essay. Do not make assumptions or incorporate external knowledge.

\vspace{1em}
\textbf{User Prompt}

Your job is to evaluate whether an essay demonstrates evidence of a particular subskill, and assign it one of five categories.

\vspace{0.5em}
\texttt{\{PROFICIENCY\_LEVEL\_DEFINITIONS\}} is inserted here. Refer to \citep{peczuh2025toward} for the general definitions of the proficiency levels used in the dataset (\textit{Not Applicable}, \textit{Below Emerging}, \textit{Emerging}, \textit{Expanding}, and \textit{Exemplifying}).

\vspace{0.5em}
You will be given a subskill and its definition, followed by the rubric corresponding to the above five categories for that subskill. Your task is to rate the given essay against the rubric and provide an explanation for the same.

\vspace{0.5em}
\textbf{Output format:}
\begin{itemize}[noitemsep, topsep=0pt]
    \item First line: \texttt{Label - <number>: <label name>}
    \item Second line: \texttt{Justification - <your justification>}
\end{itemize}

\vspace{0.5em}
\textbf{Input}

\vspace{0.5em}
\textbf{Subskill Information}\\
\texttt{Subskill - \{subskill\}}\\
\texttt{Definition - \{definition\}}

\vspace{0.5em}
\texttt{\{RUBRIC\_DESCRIPTION\}} is inserted here. Refer to \citep{peczuh2025toward} for the rubric descriptions corresponding to the above five categories for the given subskill.

\vspace{0.5em}
\textbf{Essay to be rated}\\
\texttt{\{essay\}}

\vspace{0.5em}
\textbf{Output of Essay to be rated}\\
Strictly adhere to the prescribed output format above. Do not enter any additional words, commentary, or preambles; only the two required lines.

\end{tcolorbox}
\caption{Prompt used for prompting-based baseline for rubric label generation using GPT-5-mini and GPT-5. The general proficiency-level definitions, subskill names and definitions, and the corresponding rubric descriptions are taken from \citep{peczuh2025toward}. The placeholder \texttt{\{essay\}} represents the essay to be rated.}
\label{fig:prompting-rubric-label-generation}
\end{figure*}

\begin{table*}[t]
\centering
\caption{Trait schema for the rubric-agnostic analytic traits used in our framework, including the annotation question and value scale associated with each trait. These traits capture lower-level argumentative properties that may generalize across rubrics.}
\renewcommand{\arraystretch}{1.2}
\begin{tabular}{p{3.0cm} p{4.4cm} p{8.4cm}}
\toprule
\textbf{Trait} & \textbf{Annotation Question} & \textbf{Values and Definitions} \\
\midrule

\textbf{Claim Explicitness}
& Does the essay clearly state a central, arguable position?
& \textit{Absent}: no identifiable central claim \newline
  \textit{Vague}: position is implied or unclear \newline
  \textit{Explicit}: clear, arguable central claim \\

\textbf{Structural Coherence}
& Is the reasoning logically organized and internally consistent?
& \textit{Low}: disorganized or contradictory reasoning \newline
  \textit{Medium}: mostly logical with some gaps or jumps \newline
  \textit{High}: clearly organized and internally consistent \\

\textbf{Evidence Presence}
& Does the essay provide concrete support for its claims?
& \textit{None}: no supporting examples, facts, or data \newline
  \textit{Limited}: some support but sparse or generic \newline
  \textit{Substantial}: consistent and substantive support \\

\textbf{Evidence--Claim Linkage}
& Does the essay explain why its evidence supports the claim?
& \textit{None}: no explanation linking support to claim \newline
  \textit{Partial}: weak or implicit explanation \newline
  \textit{Clear}: explicitly explains why the support justifies the claim \\

\textbf{Alternative Perspectives}
& Does the essay consider viewpoints other than its own?
& \textit{None}: no alternative perspectives acknowledged \newline
  \textit{Mentioned}: alternatives noted but not developed \newline
  \textit{Engaged}: alternatives discussed and responded to \\

\textbf{Analytical Depth}
& Does the essay go beyond description to analyze or evaluate ideas?
& \textit{Descriptive}: mostly summary or assertion \newline
  \textit{Moderate}: some reasoning or evaluation \newline
  \textit{Strong}: sustained and deeper analysis \\

\bottomrule
\end{tabular}
\label{table:traits}
\end{table*}

\begin{table}[]
\caption{Comparing different configurations of rubric and justification. Here `Desc' refers to Description and `Just' refers to Justification.}
\centering
\small
\setlength{\tabcolsep}{3pt}
\renewcommand{\arraystretch}{1.0}
\begin{tabular}{llccc}
\toprule
\textbf{Description} & \textbf{Justification} & \textbf{Acc} & \textbf{F1} & $\pmb{\alpha}$ \\
\midrule
\multirow{2}{*}{Without Desc} 
& Without Just & 0.226 & 0.158 & 0.019 \\
& With Just    & 0.284 & 0.220 & 0.007 \\
\midrule
\multirow{2}{*}{With Desc} 
& Without Just & 0.362 & 0.279 & 0.318 \\
& With Just    & \textbf{0.399} & \textbf{0.353} & \textbf{0.346} \\
\bottomrule
\end{tabular}
\label{tab:rubric_justification}
\end{table}

\subsection{Implementation Details}
\label{sec:implementation}

\paragraph{Model and Training.}
We fine-tune an open-source LLM based on Llama 3.1 8B Instruct \citep{grattafiori2024llama} using parameter-efficient fine-tuning (PEFT) with LoRA adapters \citep{hu2022lora}. We use a rank of 32, scaling factor $\alpha = 64$, and dropout of 0.05, updating only the adapter weights. Training follows the input--output formatting detailed in Section~\ref{sec:methods}, where the model takes an essay, rubric specification, and optionally trait annotations as input, and generates outputs consistent with the selected configuration (e.g., rubric labels, optionally with justifications and/or trait annotations).

We optimize using 8-bit paged AdamW optimizer \cite{loshchilov2017decoupled, dettmers2023qlora} with a learning rate of $2 \times 10^{-4}$ for 3 epochs, with a batch size of 4 and gradient accumulation of 4. We use a maximum sequence length of 1536 tokens for settings without traits and 2048 tokens for settings with traits. We select the checkpoint with the lowest validation loss on a held-out validation set across training epochs.

At test time, we generate outputs for unseen essay--rubric pairs using the same input--output formatting as during training. Generation is performed using low-temperature sampling (temperature = 0.1) with top-$k$ filtering ($k=50$), allowing up to 512 new tokens. This setting produces stable, near-deterministic outputs while retaining limited stochasticity.

All experiments are conducted on a single NVIDIA L40S GPU (48GB VRAM), with end-to-end training and inference taking approximately 6--8 hours.

\paragraph{Trait Representation Generation.}
For trait representation generation, we use GPT-4.1 with temperature 0.3, top-$p$ 1.0, and a maximum output length of 1500 tokens (see Figure~\ref{fig:trait-prompt} in the Appendix). To improve annotation reliability, we generate three outputs per essay, select the categorical trait value by majority voting, and choose the corresponding observation as the longest observation among outputs assigned to the majority value.

\paragraph{Justification Generation and Prompting-based Baselines.}
Following \citep{peczuh2025toward}, we use a similar prompt formulation for justification generation and prompting-based rubric label generation, with temperature 1.0 and top-$p$ 0.95 in both cases. For justification generation, we use GPT-4.1 with a maximum output length of 500 tokens (see Figure~\ref{fig:justification-prompt} in the Appendix). For the prompting-based baselines, we use GPT-5 with default reasoning effort and a maximum output length of 3000 tokens to support reasoning and justification generation (see Figure~\ref{fig:prompting-rubric-label-generation} in the Appendix).

\subsection{Role of Description and Justification}
\label{sec:res_desc_and_just}

Following \citep{peczuh2025toward} we consider four variants of the LLM-based framework for rubric-based essay scoring: \textit{With Description} vs.\ \textit{Without Description}, and \textit{With Justification} vs.\ \textit{Without Justification}. In the \textit{With Description} setting, the input includes the rubric description (mapping to proficiency levels), while in the \textit{Without Description} setting it is omitted and only the rubric name and definition are provided. In the \textit{With Justification} setting, the model is trained to generate both the label and a justification, whereas in the \textit{Without Justification} setting it is trained to generate only the label. These variants allow us to isolate the role of input-side rubric grounding and output-side reasoning in generalization across unseen rubrics. 

Table~\ref{tab:rubric_justification} shows results for different configurations of rubric descriptions and justifications. We observe that the setting with ``With Description'' and ``With Justification'' achieves the highest performance across all metrics. Compared to the setting with ``Without Description'' and ``Without Justification'', this configuration improves accuracy from 0.226 to 0.399 (+17.3\%), F1 from 0.158 to 0.353 (+19.5\%), and $\alpha$ from 0.019 to 0.346 (+32.7\%). Compared to the setting with ``With Description'' and ``Without Justification'', this configuration improves accuracy from 0.362 to 0.399 (+3.7\%), F1 from 0.279 to 0.353 (+7.4\%), and $\alpha$ from 0.318 to 0.346 (+2.8\%). This result suggests that descriptions provide grounding of evaluation criteria, while justifications introduce additional reasoning signals that improve alignment. Hence, we use the ``With Description'' and ``With Justification'' setting for all experiments exploring our proposed design space.

\begin{table*}[t]
\caption{Table showing case studies from the \textit{No Labels} setting where trait-output-based fine-tuning (\textit{Output: Trait Values}) predicts the correct rubric label while the corresponding \textit{No Traits} baseline does not. We include both shortened original justifications and predicted trait values to show how the predicted trait values influence the final rubric label prediction.}
\centering
\footnotesize
\renewcommand{\arraystretch}{1.15}
\begin{tabularx}{\linewidth}{p{2.6cm} p{1.45cm} X X p{4.3cm}}
\toprule
\textbf{Rubric} & \textbf{Gold} & \textbf{Output: Trait Values} & \textbf{No Traits Baseline} & \textbf{Predicted Trait Values} \\
\midrule

2.1 Synthesizing Multiple Sources
& Not Applicable
& \textbf{Not Applicable}. ``The essay does not cite any sources, making it impossible to evaluate the student's ability to synthesize multiple pieces of information.''
& \textbf{Expanding}. ``The essay cites multiple sources of information (e.g., the Mona Lisa, various forms of artistic expression) and integrates these examples to support the argument \ldots''
& Claim Explicitness: explicit \newline
  Structural Coherence: medium \newline
  Evidence Presence: limited \newline
  Evidence--Claim Linkage: partial \newline
  Alternative Perspectives: mentioned \newline
  Analytical Depth: descriptive \\
\midrule

3.1 Using Counterarguments
& Exemplifying
& \textbf{Exemplifying}. ``The essay acknowledges specific opposing viewpoints \ldots and develops responses to them by refuting the claim that online schooling eliminates bullying....demonstrating evaluation and integration of multiple perspectives \ldots''
& \textbf{Expanding}. ``The essay acknowledges specific opposing viewpoints \ldots and develops responses to them by pointing out potential negative consequences like increased cyberbullying and decreased access to resources \ldots''
& Claim Explicitness: explicit \newline
  Structural Coherence: high \newline
  Evidence Presence: substantial \newline
  Evidence--Claim Linkage: clear \newline
  Alternative Perspectives: engaged \newline
  Analytical Depth: strong \\
\midrule

4.1 Drawing Conclusions
& Emerging
& \textbf{Emerging}. ``The essay draws a specific conclusion that the Electoral College should be kept \ldots but it does not use valid assumptions or analyze complex relationships between the Electoral College and its effects.''
& \textbf{Exemplifying}.``The essay draws a specific conclusion that the Electoral College should be kept \ldots and supports this conclusion with evidence from sources, demonstrating the use of valid assumptions to analyze complex relationships \ldots''
& Claim Explicitness: explicit \newline
  Structural Coherence: medium \newline
  Evidence Presence: limited \newline
  Evidence--Claim Linkage: partial \newline
  Alternative Perspectives: none \newline
  Analytical Depth: descriptive \\
\bottomrule
\end{tabularx}
\label{tab:trait_case_studies}
\end{table*}

\subsection{Qualitative Case Studies Where Trait Outputs Help}
\label{sec:case_studies}
Table~\ref{tab:trait_case_studies} shows three examples from the \textit{No Labels} setting where trait-output-based fine-tuning (\textit{Output: Trait Values}) predicts the correct rubric label while the corresponding \textit{No Traits} baseline does not. These examples help explain why trait outputs are useful in the hardest setting: they provide intermediate signals that guide the model toward rubric decisions that are better aligned with the essay.

\paragraph{Example 1: Rubric 2.1 (Synthesizing Multiple Sources).}
Rubric 2.1 evaluates whether the essay \emph{effectively synthesizes multiple pieces of information}. In this example, for essay ID \texttt{20D57DEC3EF5}, the trait-output model correctly predicts \emph{Not Applicable}, while the baseline overpredicts \emph{Expanding}. At a high level, the trait-output justification recognizes that the essay does not provide the source-based support needed to evaluate synthesis, whereas the baseline incorrectly interprets the essay as integrating multiple sources. The predicted traits help explain this correction: \emph{limited} \textbf{Evidence Presence} suggests that the essay lacks sufficient supporting material for cross-source synthesis, and \emph{descriptive} \textbf{Analytical Depth} indicates that the essay does not reason over multiple pieces of information in the way this rubric requires. Together, these traits make the lower label more appropriate.

\paragraph{Example 2: Rubric 3.1 (Using Counterarguments).}
Rubric 3.1 evaluates whether the essay \emph{effectively addresses counterarguments}. In this example, for essay ID \texttt{F0AA6B265B6F}, the trait-output model correctly predicts \emph{Exemplifying}, whereas the baseline predicts the lower label \emph{Expanding}. Both outputs recognize that the essay acknowledges opposing views, but the trait-output justification better captures that the essay not only mentions counterarguments but also responds to them in a developed way. This stronger prediction is supported by two key traits: \emph{engaged} \textbf{Alternative Perspectives} shows that opposing viewpoints are actively discussed and addressed rather than merely noted, while \emph{strong} \textbf{Analytical Depth} indicates sustained reasoning about those viewpoints. These intermediate signals are well aligned with the highest label for this rubric.

\paragraph{Example 3: Rubric 4.1 (Drawing Conclusions).}
Rubric 4.1 evaluates whether the essay \emph{draws specific conclusions}. In this example, for essay ID \texttt{845DB3AC386E}, the trait-output model correctly predicts \emph{Emerging}, while the baseline overpredicts \emph{Exemplifying}. Both justifications identify that the essay reaches a conclusion, but the trait-output model assigns a more moderate label that better matches the gold score. The relevant trait signals are \emph{medium} \textbf{Structural Coherence} and \emph{partial} \textbf{Evidence--Claim Linkage}. These signals suggest that, although the essay does arrive at a conclusion, the reasoning is not fully developed and the support is not fully connected back to the claim. As a result, the essay is better characterized as showing a conclusion in an emerging form rather than as demonstrating the stronger reasoning expected for a top label.

Overall, these examples suggest that trait outputs help by making rubric-relevant argumentative properties explicit. In Examples 1 and 3, they help prevent overly strong predictions by revealing missing support or incomplete reasoning. In Example 2, they help recover the highest label by highlighting deep engagement with counterarguments. This qualitative pattern is consistent with the broader quantitative finding that trait-output-based fine-tuning is especially helpful in the \textit{No Labels} setting.

\subsection{Limitations}

Some limitations of our work include the use of automatically generated trait annotations and the scope of our empirical evaluation. First, the trait annotations used in our framework are generated using LLM prompting and are not validated through human annotation, unlike the gold rubric labels in the dataset. We therefore treat these trait annotations as silver intermediate structure that supports model training, rather than as fully reliable standalone supervision targets. Although our results suggest the value of such trait-based structure for improving cross-rubric generalization, especially in low-supervision settings, future work could examine annotation quality more directly through human evaluation or stronger validation procedures.

Second, our empirical study focuses on one recent benchmark for critical thinking essay scoring and on a Llama-based open-source model within our experimental setup. This setting provides a focused testbed for studying cross-rubric generalization. Future work may help assess the extent to which the reported findings generalize to other datasets, writing domains, rubric designs, and model families.

\subsection{Ethical considerations}

Our work studies automated scoring of student essays, which requires careful use in educational settings. First, the trait annotations in our framework are automatically generated using LLM prompting rather than validated through human annotation. We use them as intermediate structure to support training, but they may still reflect model errors or biases.

Second, although automated rubric-based scoring may be useful for scalable assessment support, it should not be treated as a replacement for human judgment in high-stakes educational decisions. In addition, parts of our pipeline rely on proprietary APIs, so any real-world use would require appropriate safeguards for student data.

Finally, our experiments require GPU compute for fine-tuning and inference, which may have some environmental impact.

%
%

\end{document}